\pdfoutput=1

\documentclass[11pt]{article}

\usepackage[final]{acl}

\usepackage{times}
\usepackage{latexsym}
\usepackage{multirow}
\usepackage{framed}

\usepackage[T1]{fontenc}

\usepackage[utf8]{inputenc}

\usepackage{microtype}

\usepackage{inconsolata}

\usepackage{graphicx}

\title{Rehearse With User: Personalized Opinion Summarization via Role-Playing based on Large Language Models}

\author{\bf{Yanyue Zhang}$^{\spadesuit}$, \bf{Yulan He}$^{\heartsuit}$ and \bf{Deyu Zhou\thanks{\;Corresponding author.}}$^{\spadesuit}$ \\
$^{\spadesuit}$School of Computer Science and Engineering, Key Laboratory of Computer Network \\ and Information Integration, Ministry of Education, Southeast University, China \\
$^{\heartsuit}$Department of Informatics, King’s College London
$^{\heartsuit}$The Alan Turing Institute \\
\texttt{\{yanyuez98,d.zhou\}@seu.edu.cn},\\
  \texttt{yulan.he@kcl.ac.uk}}

\begin{document}
\maketitle

\begin{abstract}
Personalized opinion summarization is crucial as it considers individual user interests while generating product summaries.
Recent studies show that although large language models demonstrate powerful text summarization and evaluation capabilities without the need for training data, they face difficulties in personalized tasks involving long texts. 
To address this, \textbf{Rehearsal}, a personalized opinion summarization framework via 
LLMs-based role-playing is proposed. 
Having the model act as the user, the model can better understand the user's personalized needs.
Additionally, a role-playing supervisor and practice process are introduced to improve the role-playing ability of the LLMs, leading to a better expression of user needs.
Furthermore, through suggestions from virtual users, the summary generation is intervened, ensuring that the generated summary includes information of interest to the user, thus achieving personalized summary generation.  
Experiment results demonstrate that our method can effectively improve the level of personalization in large model-generated summaries.
\end{abstract}

\section{Introduction}



Personalized opinion summarization, which takes into account user characteristics and interests while summarizing multiple product reviews, aims to meet the individual needs of users. Based on general multi-document opinion summarization, personalized opinion summarization needs to understand user preferences from relevant historical information and analyze aspects of the current product that the user may be interested in. Then, based on the user's interests, it provides a targeted summary of product reviews, generating more content that the user is interested in. Due to the difficulty in annotation, research related to personalized opinion summarization is almost nonexistent.


\begin{figure*}[t]
\centering
\includegraphics[width=0.95\textwidth]{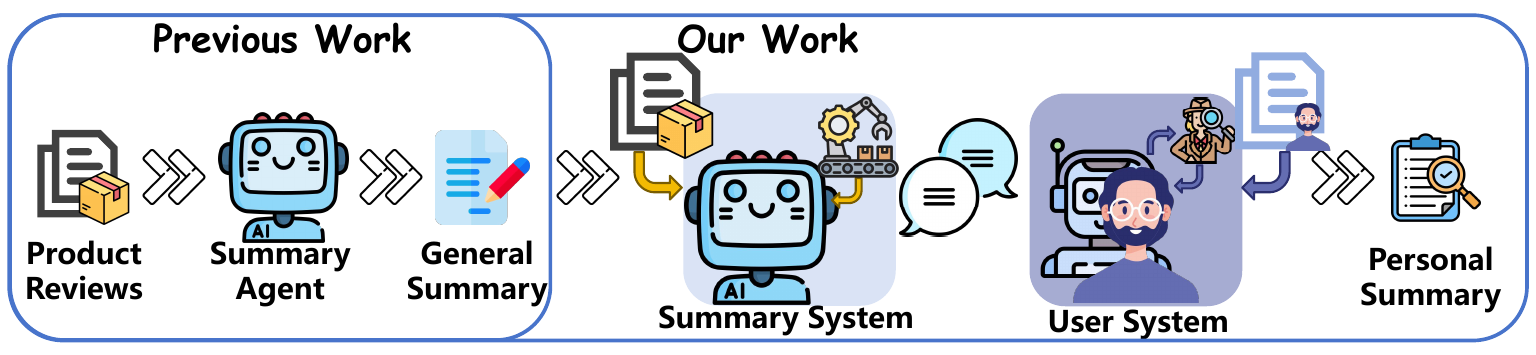} 
\caption{The difference between previous work and our work, \textbf{Rehearsal}. The summarization system primarily inputs the product review set and is enhanced via the retriever. The user system primarily inputs the user review set and is enhanced using a supervisor.}
\label{fig1}
\vspace{-0.2cm}
\end{figure*}

Recent studies have shown that large language models (LLMs), such as GPT-3.5, can achieve state-of-the-art, even human-level, performance on standard summarization benchmarks without the need for large-scale training data \citep{korkankar-etal-2024-aspect,pu2023summarization,wang2024iteratively}. Moreover, some works have demonstrated that LLM-based summary evaluation metrics can more flexibly assess different aspects of summary generation, showing stronger correlations with human evaluation, compared to traditional metrics \citep{song2024finesure,siledar-etal-2024-one}. However, these works are mostly limited to generic summarization scenarios.
In scenarios like personalized recommendations\citep{li2023personalized,li2023prompt,chen2022personalized,yang2023palr}, search\citep{baek2024knowledge,salemi2024towards}, and dialogue\citep{hudecek-dusek-2023-large,hu2023enhancing,yang-etal-2023-refgpt}, some studies have shown that LLMs have certain personalization capabilities.

However, it has also been found that personalized LLMs also face difficulties in long-text scenarios\citet{richardson2023integrating,tseng2024two}.
On the one hand, handling long texts with extensive redundant information is challenging for LLMs\citep{nayab2024concise,shi2023large}. Excessively long input texts easily exceed the model's input length limit. Besides, there is usually a lot of redundant information from product reviews or user history in multi-document opinion summarization, which greatly hinders LLMs from understanding product features and users' interests accurately.

On the other hand, it is highly difficult to infer users' preferences for the current product from their complex historical data \citet{richardson2023integrating,tseng2024two}. Users' shopping preferences are not explicit but hidden within a large volume of historical information. Moreover, their interest in the current product cannot be directly equated to their preferences for other products, requiring intricate analysis.

Therefore, we propose \textbf{Rehearsal}, a personalized opinion summarization framework based on LLMs role-playing. As shown in Figure \ref{fig1}, To alleviate the input pressure on the LLMs from complex user historical information and a large volume of product review texts, we adopt a multi-agent framework, which models product information and user information separately through the summary system and the user system. Via the interaction between these two agent systems, the personal summary is generated. To enable the user system to better understand user interests, role-playing with supervision and practice is introduced into the user system. 

Specifically, the framework consists of three steps: generic summary generation, user suggestions based on role-playing with supervision and practice, and retrieval-augmented personalized summary generation. First, we generate a generic opinion summary based on product reviews. Second, an LLM acts as the user and proposes modifications to the current summary. To ensure consistency in role-playing, a professional role-playing observer is introduced. The observer provides continuous modification suggestions based on four dimensions of user consistency, ensuring that the user model remains true to the role. Before formal role-playing, an exercise process is executed in advance, where the observer conducts role-playing practice with the user model. The results of the process are then used to strengthen the user model in the formal role-playing. Third, after receiving user suggestions, retrieval augmentation is applied to eliminate irrelevant product reviews and extract important user reviews. The summary is then revised based on the retrieved text and the suggestion.

Our main contributions are as follows:

\begin{itemize}
\item We explore LLMs-based personalized opinion summarization generation and evaluation.
\item We improve the personality ability of LLMs through user role-playing based on supervision and practice.
\item Experiments have proven that the summaries generated by Rehearsal are more aligned with users' personalized needs.
\end{itemize}

\section{Related Work}

\subsection{Personalized Opinion Summarization}

Opinion summarization\citep{chu2019meansum,copycat-bravzinskas2020unsupervised,dae-amplayo2020unsupervised,iso2021convex,zhang2023disentangling} generally focuses on user reviews about products, hotels, restaurants, and so on. 
Due to the challenges of annotation and evaluation of personalized opinion summarization, there has been no research on this topic. Previous efforts have focused on simplifying the problem into either controllable summarization or user-interaction-based summarization~\citep{zhang-etal-2024-opinions,zhang-zhou-2023-disentangling,hosking2023attributable,CARICHON2024124449,SYED2024100238,BENEDETTO2024124567,yan2011summarize,zhang2025personalsum}. 
Other works have explored personalized review summarization in single-document settings, where titles written by users serve as summaries for the corresponding review texts~\citep{xu2023pre,cheng2023towards,xu2023sentiment}. However, those methods aim to model the personalization of the review authors, while personalized opinion summarization in this study focuses on understanding the preferences of readers engaging with multiple reviews.

\subsection{Role playing and Multi-agent based on LLMs}
In recent years, Large Language Models have demonstrated significant potential in reasoning and planning capabilities, aligning perfectly with human expectations for autonomous agents that can perceive their surroundings, make decisions, and take actions accordingly\citep{xi2025rise,wooldridge1995lntelligent,russell2016artificial,guo2023can,liang2023let}. Building on this, some studies have proposed LLM-based multi-agent systems\citep{guo2024large}, leveraging the collective intelligence\citep{liu2023dynamic,hongmetagpt}, specialized roles\citep{li2023camel,dong2024self}, and interactions of multiple agents based on the powerful capabilities of a single LLM agent \citep{Duimproving,xiong2023examining,chanchateval,mao2023alympics,mandi2024roco,hongmetagpt}.


\subsection{LLMs Personalization}

Research related to personalized LLMs primarily focuses on how to meet user expectations and fulfill their needs. To enhance individual preferences, personalized LLMs consider user personas (e.g., personal information, historical behaviors) and cater to customized needs\citep{chen2024large,personalize-2024-personalization}. 
To enhance individual preferences, some studies explore various instruction and framework designs\citep{li2023personalized,li2023prompt,yang-etal-2023-refgpt,li2024guiding,hu2023enhancing}, while others focus on fine-tuning model parameters to better understand the personalized demands of special tasks\citep{chen2022personalized,yang2023palr,hudecek-dusek-2023-large}. 

\citet{richardson2023integrating,tseng2024two} point out that incorporating user history data into the prompt to personalize LLMs could lead to input exceeding context length and increase inference costs.
Unlike other studies that focus purely on historical information retrieval \citep{richardson2023integrating,zhong2024memorybank,zhang2024personalized,sun2024persona}, this paper enhances the LLM's ability to understand personalized needs by employing user role-playing.

\section{Methodology}

In this section, we provide a detailed introduction to our method for personalized opinion summarization based on user role-playing. Given a set of reviews about an entity (e.g., a product) and a collection of historical comments from a specific user, the aim is to summarize the opinions that are of interest to the user, as expressed in the product reviews. Next, we will introduce the overall process of the method. In the subsequent sections, we will explain the three components of the method: general summary generation, user suggestions based on practice and supervision, and how to perform retrieval-augmented personalized summary rewriting based on personalized suggestions.

\begin{figure*}[!t]
\centering
\includegraphics[width=0.90\textwidth]{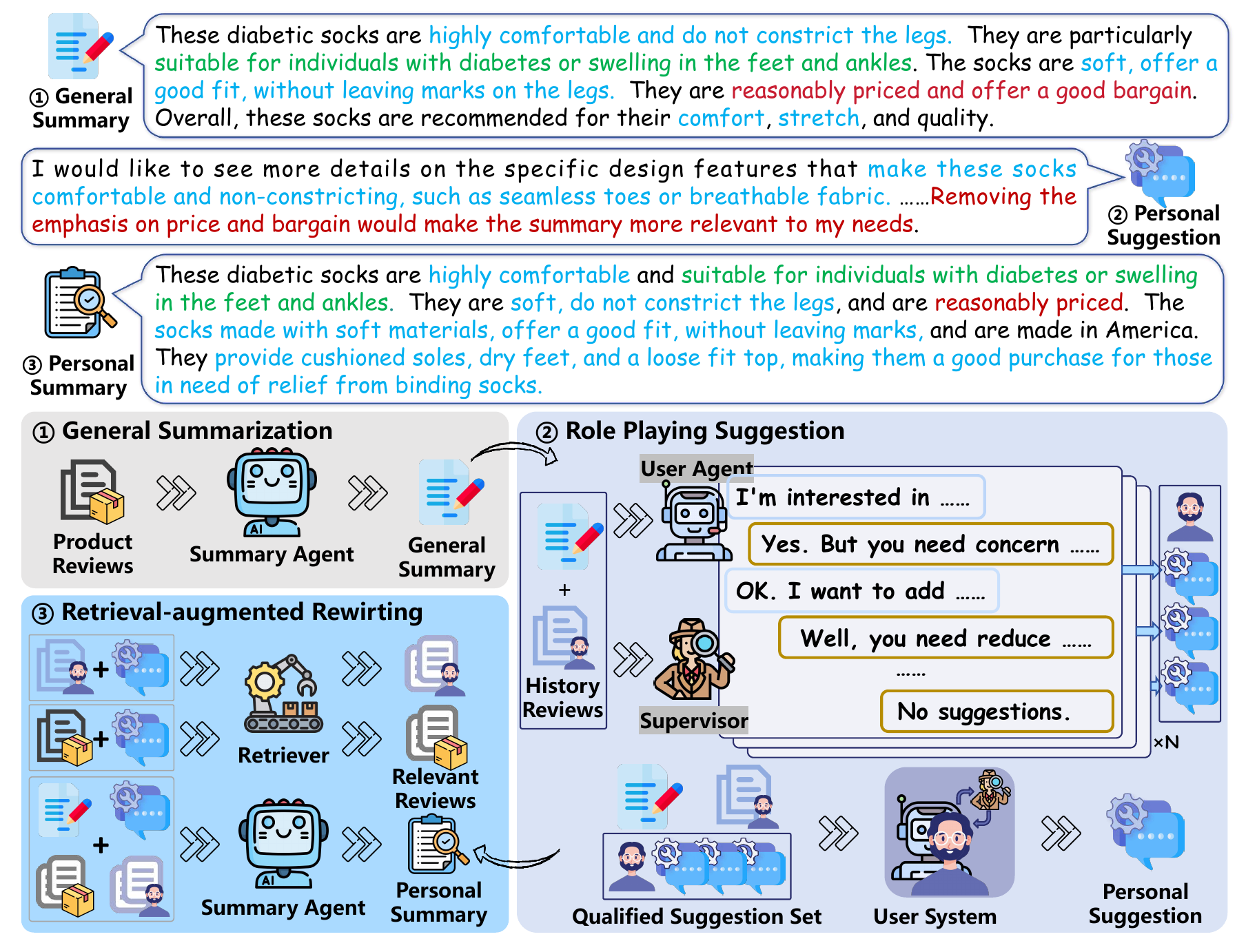} 
\caption{The example (up) and the execution process (below) of \textbf{Rehearsal}. The example includes the output of three steps. Different colors represent different aspects of the product information. The LLMs-based pseudo-user suggests the summary agent add comfort-related (blue) information and reduce price-related (red) information.}
\label{fig2}
\vspace{-0.2cm}
\end{figure*}

\subsection{Overview}

As shown in Figure \ref{fig2}, To alleviate the pressure of input length on the model's understanding ability, Rehearsal follows a "generate first, rewrite later" architecture to achieve personalized summarization. The method mainly involves three types of models: (1) a summary agent that processes product reviews and generates general summaries, (2) a user agent and role-playing supervisor responsible for modeling user information and generating and evaluating personalized opinions, and (3) a historical filter and a product filter that primarily filter product reviews and historical comments based on suggestions, respectively.

Specifically, the method is divided into three parts: general summarization, role-playing suggestions with practice and supervision, and retrieval-augmented personalized summary rewriting. In the first part, the summary agent summarizes product reviews to generate a general summary. Then, the user agent, based on the user's historical comment information, role-plays the user and provides modification suggestions for the currently generated summary. During this stage, the user agent first undergoes role-playing practice. Under the guidance of a professional supervisor, the user agent continuously attempts to generate suggestions. After the generated suggestions are evaluated by the supervisor, the user agent will generate and modify the personalized suggestions based on the suggestions accumulated during the practice phase until they pass the supervisor's evaluation. Finally, retrieval-augmented rewriting is performed. Since rewriting the summary only requires information relevant to the modification suggestions, product and historical reviews are filtered based on the suggestions. By selecting review texts related to the suggestions, redundant information that could affect the rewriting process and unnecessary inference overhead are avoided. Ultimately, under the guidance of the modification suggestions provided by the user agent, the summary agent revises the summary based on the relevant product reviews and historical comments, resulting in the final personalized summary.

\subsection{General Summarization}
Due to the performance of LLMs being highly susceptible to interference from text length, positional biases, and irrelevant information \citep{shi2023large,nayab2024concise}, the method follows a framework of first generating a general summary and then performing personalized rewriting. In the general summary generation phase, the summary agent will initially summarize the product reviews. The prompt used adopts the simplest structure, including instructions and output format specifications. The instruction text is: "Your task is to generate a summary of the current product review." The output format specification is provided in the form of a dictionary.

\subsection{Role-Playing Suggestions}

The user role-playing suggestions mainly rely on the interaction between the user agent and the role-playing supervisor. It consists of two phases: the practice phase and the formal suggestion phase. Specifically, the role-playing supervisor evaluates the user's response rationality from four aspects: historical exposure rate, knowledge accuracy, knowledge hallucination rate, and personal utterance consistency\citep{tu2024charactereval} and provides modification suggestions. 

In the practice phase, the process simulates the formal suggestion procedure by generating qualified personalized suggestions through the interaction of the two models. The user agent needs to adjust its responses based on the supervisor's suggestions until the supervisor determines that the generated text aligns with role consistency and no further suggestions are required. These qualified personalized suggestions will serve as examples for the formal suggestion phase, helping the user model better understand the user's characteristics. In the formal generation process, the user agent, with the assistance of the role-playing supervisor, will generate and modify suggestions until the generated suggestions pass the supervisor’s evaluation.

\subsubsection{User Agent Design}

The User-Agent primarily understands the user's interests and generates suggestions based on the user's historical reviews. To help the user agent better understand user interests, the agent is required to first generate an analysis of "self" and the current summary, followed by generating suggestion text. The self-analysis includes a self-introduction, mainly containing information about the product aspects the user is interested in or not interested in. The summary analysis identifies which parts of the current summary align with the user’s interests and which do not and what additional product information the user would like to know. The suggestion text should be concise modification advice, mainly focusing on which product aspects should be enhanced or reduced.

The instructions for the user agent are divided into four parts: task description, thought chain, notes, and output format specification. The task description is: "Your task is to act as the user based on their historical reviews, evaluating whether the current summary addresses aspects you are interested in, and providing suggestions for modifications to the summary of the current product review." In the thought chain and output format specification, the content and form of the user agent's response are constrained. The notes emphasize that the current task is role-playing, and the model should reply in the first person.
The complete prompt content can be found in the Appendix \ref{user_prompt}.

\subsubsection{Role-Playing Supervisor Design}
To allow the role-playing supervisor to comprehensively evaluate the user agent's consistency with role-playing, following the work of A study, we introduce four metrics: historical exposure rate, knowledge accuracy, knowledge hallucination rate, and personal utterance consistency. Among these, the first three belong to Knowledge Consistency, and the last one belongs to Persona Consistency.
\textbf{Historical exposure rate} refers to the amount of character-related knowledge or information present in the response.
\textit{Knowledge accuracy} refers to the correctness of the knowledge or information utilized in the response.
\textit{Knowledge hallucination} rate refers to the extent to which a response contains inappropriate information that the character should not.
\textit{Personal utterance consistency} refers to the consistency in a response with the character’s personality and speech habits.

To further assist the supervisor in responding to the user's reply, the thought chain in the instructions details the evaluation process for the supervisor, including analyzing inputs, evaluating each instruction, and summarizing outputs.
(1) In the analysis section, the LLM is required to understand the user's perception of self, the summary, and the suggestions based on the response generated by the user agent. Then, the supervisor will further analyze the aspects and sentiment of the product being discussed based on the general summary. Finally, the supervisor will understand the user's personality, shopping behavior, and interests from the user's historical comments.
(2) In the evaluation section, The evaluation points for the four metrics are further described to guide the model in assessing these metrics.
(3) In the Summarization section, the observer is required to first summarize the previous evaluation results and generate brief reasons and clear suggestions for any errors found.

The instruction design consists of five parts: task description, notes, metric introduction, thought chain, and output format. The task description is: 
"You are a role review expert, skilled in identifying and correcting any anomalous text in dialogue that may not align with the user's personality. Your goal is to evaluate whether the user's response is consistent with their previous behavior based on historical comments, and to offer improvement suggestions. The suggested content should include a brief reason and specific, detailed revision advice. IF THERE ARE NO ERRORS OR SUGGESTIONS, you must write ONLY 'No suggestions' in the suggestions section, without any explanation or additional words."
The complete prompt content can be found in the Appendix \ref{super_prompt}.

\subsection{Retrieval-augmented Personalized Rewriting}

Since rewriting the summary does not require re-browsing all the product reviews but only focuses on the portions relevant to the modification suggestions, the product reviews are filtered based on the suggestions to eliminate irrelevant text. Moreover, to strengthen the personalization of user information, texts that are more relevant to the current product, based on the user's historical reviews and modification suggestions, are selected as additional personalized suggestions. Finally, the summary agent modifies the previously generated general summary based on the filtered product reviews, personalized suggestions, and related historical reviews to generate a personalized summary. The instruction design for the summary agent and the two filters includes a task description, notes, and output format specification. Specific prompts are provided in the Appendix \ref{rag-prompt}.

\section{Experiments} 

\subsection{Datasets}
Due to the lack of research on personalized opinion summarization, we have constructed PerSum, a personalized opinion summarization test set based on the Amazon dataset \citep{copycat-bravzinskas2020unsupervised}. The data includes four categories: Clothing, Shoes and Jewelry, Electronics, Health and Personal Care, and Home and Kitchen, with a total of 666 samples. Each sample contains a set of product reviews, a set of the user's historical reviews on products in the same category, and the user's review of the current product.

To ensure the quality of the dataset, three rounds of filtering were applied to the original Amazon data. First, the number and length of the user’s historical reviews were filtered. Only samples with more than five but fewer than 50 reviews and a total review length under 27,000 characters are retained. Second, the personalized review was evaluated using ROUGE-\{1, 2, L\} metrics \citep{lin2004rouge} against the historical and product review sets and samples with a total score of ROUGE-\{1, 2, L\} below 0.45 were removed to ensure the quality of the personalized reviews. 

Finally, aspect coverage and sentiment consistency scores from OP-I-MISTRAL \cite{siledar-etal-2024-one} were introduced to further evaluate the personalized reviews. Samples with both aspect and sentiment scores higher than 4 for personalized reviews and product reviews were categorized as high product-scoring samples, while those with scores lower by one were categorized as low-scoring samples. For each product category, a certain number of samples were extracted, with both the historical and product reviews scoring high and others scoring low. Additionally, to increase the difficulty of personalization, some high-scoring historical samples with low product scores were added. These samples' personalized reviews have a higher relevance to the user’s historical reviews but differ from those of other users for the current product. More analysis is provided in the Appendix \ref{dataset}.

\subsection{Evaluation Metrics and Baselines}

\begingroup
\renewcommand{\arraystretch}{1.10}
\begin{table*}[!th]
\centering
\scalebox{0.88}{
\begin{tabular}{ll|cc|cc|cc|c}
\hline \multirow{2}{*}{\textbf{Model}} & \multirow{2}{*}{\textbf{Method}} & \multicolumn{2}{c|}{\textit{Product}} & \multicolumn{2}{c|}{\textit{History}} & \multicolumn{2}{c|}{\textit{User}} & \multirow{2}{*}{ \textbf{AVG} } \\
 & & \textbf{AC} & \textbf{SC} & \textbf{AC} & \textbf{SC} & \textbf{AC} & \textbf{SC} & \\
\hline \multirow{5}{*}{GPT-3.5-turbo}
 & PerSum & 82.19 & 78. 14 & 74.00 & 71.99 & 73.93 & 66.32 & 74.43  \\
 & Ana+PerSum & 82.63 & 77.91 & 83.81 & 80.85 & 80.26 & 72.75 & 79.70  \\
  & OnlySum & 88.43 & 82.66 & 79.18 & 75.83 & 77.77 & 69.27 & 78.86  \\
 & Sum+PerChan &82.56 &77.94 &88.59 &86.31 &80.51 &75.41 &81.89 \\
 & Rehearsal & \textbf{91.87} & \textbf{86.41} & \textbf{88.92} & \textbf{85.34} & \textbf{89.78} & \textbf{81.60} & \textbf{87.32} \\
\hline \multirow{5}{*}{GPT-4o} 
 & PerSum & 92.88 & 88.73 & 86.57 & 82.45 & 86.75 & 76.66 & 85.67 \\ 
 & Ana+PerSum & 92.40 & 87.81 & \textbf{93.22} & \textbf{89.49}& 93.40 & 84.67 & 90.16  \\
 & OnlySum & 91.64 & 86.00 & 87.50 & 82.35 & 89.60 & 80.15 & 84.87  \\
 &Sum+PerChan&88.91 &84.52 &90.83 &88.10 &93.68 &88.22 &89.04 \\
 & Rehearsal & \textbf{94.05} & \textbf{88.45} & 92.58 & 87.19 & \textbf{96.46} & \textbf{87.82} & \textbf{91.07} \\
\hline \multirow{5}{*}{GPT-4-turbo} 
 & PerSum & 90.33 & 85.98 & 84.38 & 80.99 & 83.41 & 73.99 & 83.18  \\
 & Ana+PerSum & 92.03 & 86.92 & 92.15 & \textbf{88.54} & 92.74 & 83.62 & 89.34  \\
 & OnlySum & 88.53 & 84.46 & 81.62 & 78.00 &  78.82 & 69.78 & 80.20  \\
 &Sum+PerChan&87.27 &83.54 &84.62 &82.22 &82.89 &76.39 &82.82 \\
 & Rehearsal &\textbf{94.37} &\textbf{88.19} &\textbf{92.66} &86.51 &\textbf{96.26} &\textbf{86.82} &\textbf{90.80} \\
\hline \multirow{5}{*}{CharacterGLM-4} 
 & PerSum & 92.19 & 87.10 & 85.86 & 82.48 & 85.57 & 76.13 & 84.89  \\
 & Ana+PerSum & 89.44 & 85.24 & 90.93 & 88.13 & 91.59 & 83.91& 88.21 \\

 & OnlySum & \textbf{93.52} & \textbf{88.14} & 88.96 & 83.71 &  91.42 & 81.07 & 87.80 \\
 &Sum+PerChan&90.36 &85.42 &88.43 &86.27 &89.54 &82.36 &87.07 \\
  & Rehearsal & 92.27 & 87.67 & \textbf{92.84} & \textbf{89.35} & \textbf{93.56} & \textbf{85.98} & \textbf{90.28}
\\ 	  	  

\hline \multirow{5}{*}{Qwen-turbo} 
 & PerSum & 88.53 & 84.46 & 81.62 & 78.00 & 78.82 & 69.78 & 80.20  \\
 & Ana+PerSum & 92.65 &\textbf{ 88.62} & 92.35 & 89.24 & 94.13 & 86.19 & 90.53  \\
 & OnlySum & 93.30 & 87.96 & 88.94 & 85.06 &  91.70 & 81.72 & 88.13  \\
 &Sum+PerChan&92.58 &89.18 &91.59 &88.97 &92.87 &87.99 &90.53 \\
 & Rehearsal & \textbf{93.48} & 87.97 & \textbf{93.79} & \textbf{89.82} & \textbf{94.98} & \textbf{87.45} & \textbf{91.25} \\
\hline
\end{tabular}
}
\caption{Results from experiments for different models on PerSum. The bold scores denote the best scores.
}
\label{result1_all}
\vspace{-0.2cm}
\end{table*}
\endgroup

We use evaluation metrics based on LLMs, including aspect coverage and sentiment consistency from OP-I-MISTRAL \citep{siledar-etal-2024-one}, as the evaluation metric. For the generated personalized summaries, we measure the extent to which the summary captures product information by calculating its aspect coverage and sentiment consistency with the product review set. Similarly, we calculate two scores between the summary and historical reviews, as well as the personalized review, to evaluate how well the summary aligns with the user's interests. Experiments have shown that the average score of completely irrelevant texts exceeds 2. Therefore, we expanded the scoring range from 1-5 to 0-100, and the prompt was adjusted to evaluate at a more granular sentence level. The specific prompt can be found in the Appendix \ref{eval-metric}.

To evaluate the effectiveness of the framework we designed, we have utilized five models that are readily accessible through public APIs, including GPT-3.5, GPT-4o, GPT-4\citep{achiam2023gpt}, CharacterGLM\citep{zhou2023characterglm}, and Qwen\citep{qwen}. Our experiments include the following baselines:

\textbf{PerSum}: Inputs both the product review set and the user review set into the model, with instructions for personalized summary generation.

\textbf{Ana+PerSum}: First, the model is instructed to analyze the user's interests based on the user review set and then generate a personalized summary based on the user's interests and the product review set.

\textbf{OnlySum}: Generates a general summary using only the product reviews. The model will not receive any personalized instruction or input from the user's historical reviews.

\textbf{Sum+PerChan}: Builds on the general summary generated by \textbf{OnlySum} and asks the model to modify the summary based on the user's historical reviews to make it personalized.

\subsection{Implementation Details}
All summary generation experiments were carried out via API. For the GPT series, we used GPT-3.5-turbo, GPT-4o, and GPT-4-turbo. For CharacterGLM, we used GLM-4-Air-0111. For Qwen, we used qwen-turbo. Apart from using the results generated during the training process as demonstrations in the user role-playing, no output examples were included in any prompts, only output format instructions. For each summary generation experiment, the same LLM was used throughout the process, including the summary generator, user model, supervisor model, extractor, and models in the relevant baseline methods.

In the role-playing suggestion process, all suggestions that pass the check will be used as demonstrations in the formal suggestion stage. If all suggestions in the practice rounds for a sample fail, they will all be used to avoid the model being affected by a single erroneous example.
In the Retrieval-augmented Rewriting process, different filtering methods are applied to the product and historical review sets. For product reviews, the total character count of the review set is required to be less than 10,000. For historical reviews, only the most relevant historical review is chosen. More details can be found in Appendix \ref{detail}.

\begingroup
\renewcommand{\arraystretch}{1.10}
\begin{table*}[!th]
\centering
\scalebox{0.88}{
\begin{tabular}{ll|cc|cc|cc|c}
\hline \multirow{2}{*}{\textbf{Model}} & \multirow{2}{*}{\textbf{Method}} & \multicolumn{2}{c|}{\textit{Product}} & \multicolumn{2}{c|}{\textit{History}} & \multicolumn{2}{c|}{\textit{User}} & \multirow{2}{*}{ \textbf{AVG} } \\
 & & \textbf{AC} & \textbf{SC} & \textbf{AC} & \textbf{SC} & \textbf{AC} & \textbf{SC} & \\
\hline \multirow{6}{*}{GPT-3.5-turbo}  
 & Sum+PerChan & 82.56 & 77.94 & 88.59 & 86.31 & 80.51 & 75.41 & 81.89\\
 & \quad +RoleChange & 90.32 & 85.54 & 83.24 & 79.53 & 81.79 & 72.66 & 82.18 \\
 & \quad+Retriever & \textbf{92.98} & \textbf{87.07} & 88.64 & 84.69 & 86.74 & 77.43 & 86.26 \\
 & \quad+Supervisor & 92.05 & 86.96 & 84.94 & 81.51 & 83.98 & 75.19 & 84.11 \\
 & \quad +ICL-Super & 92.64 & 87.57 & 86.80 & 82.80 & 86.16 & 76.58 & 85.43 \\
 & ALL & 91.87 & 86.41 & \textbf{88.92} & \textbf{85.34} & \textbf{89.78} & \textbf{81.60} & \textbf{87.32} \\
\hline \multirow{6}{*}{GPT-4o} & Sum+ PerChan& 88.91 & 84.52 & 90.83 & 88.10 & 93.68 & 88.22 & 89.04 \\
& \quad +RoleChange & 93.70 & 88.71 & 91.12 & 85.63 & 93.73 & 83.01 & 89.32 \\
 &  \quad +Retriever & \textbf{94.34} & 88.90 & \textbf{92.64} & \textbf{87.54} & 95.95 & 86.63 & 91.00 \\
 &  \quad +Supervisor & 93.86 & 91.07 & 91.77 & 87.40 & 92.76 & 86.93 & 90.63 \\
 &  \quad +ICL-Super & 93.60 & \textbf{89.84} & 91.33 & 86.58 & 92.33 & 86.48 & 90.03 \\
 &  ALL & 94.07 & 88.33 & 92.25 & 87.49 & \textbf{96.46} & \textbf{87.82} & \textbf{91.07} \\
\hline
\end{tabular}
}
\caption{Ablation results for different models on PerSum. The bold scores denote the best scores.
}
\label{result_ab}
\vspace{-0.2cm}
\end{table*}
\endgroup

\begin{figure*}[!t]
\centering
\includegraphics[width=1.0\textwidth]{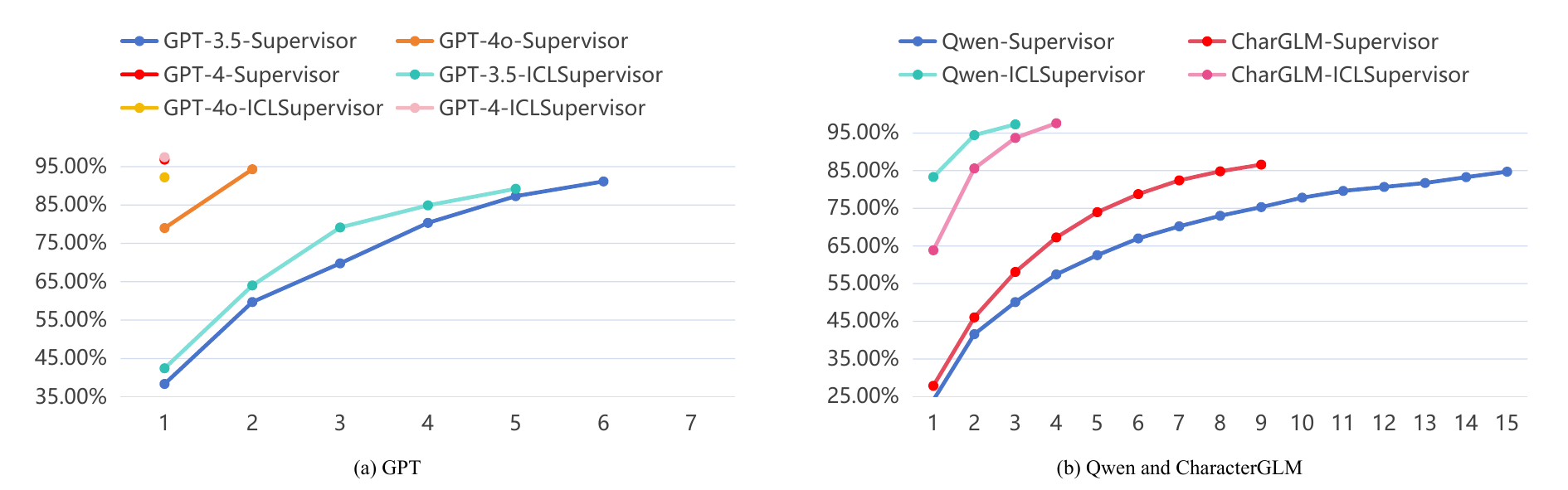} 
\caption{Relationship between pass rate and iteration count for different models, with and without ICL in user role-playing.}
\label{succ}
\vspace{-0.2cm}
\end{figure*}

\subsection{Results}
As shown in Table \ref{result1_all}, our framework Rehearsal outperforms all baseline methods in terms of overall scores across the five base models, especially the User-base scores and average scores, which are higher than all baselines. This indicates that Rehearsal enhances the personalization of the summaries while maintaining the general summarization capabilities. Specifically, Rehearsal shows significant improvements over the two-stage methods, Ana+PerSum and Sum+PerChan, when compared to the User-base scores, demonstrating that the performance improvement is not solely due to using multiple agents for modeling user and product reviews separately.

A comparison between the results of OnlySum and PerSum shows that when the input length is too long, the summary scores generated by all base models significantly decrease, both in product-related and user-related metrics. Even if the length does not exceed the limit, large amounts of information put substantial pressure on the model's generation performance. Not only does the personalization performance decline, but the general summary generation capability is also affected.

\subsection{Analysis}

To evaluate the impact of each improvement on model performance, we conducted fine-grained ablation experiments on GPT-3.5 and GPT-4o. Overall, the addition of each component design to the baseline and Sum+PerChan methods led to certain performance improvements. Among them, +RoleChange introduces only the user role-playing suggestion modification step without using RAG or the supervisor. This method shows some improvement over the baseline, Sum+PerChan, indicating that introducing user role-playing helps the LLMs better understand user interests than directly performing personalized modifications. When comparing RAG with supervisor-based user role-playing enhancement, RAG shows a more noticeable improvement in summary generation. Additionally, comparing the results of GPT-3.5 and GPT-4o reveals that the performance improvement brought by ICL-enhanced user role-playing is not stable.

To further explore the impact of using In-Context Learning (ICL) on LLMs role-playing, we compared the pass rates of LLMs role-playing with and without ICL. It is evident that after using ICL, the pass rates of all base models in role-playing improve with the increase in iteration count. The most notable improvement was observed in the Qwen model, which required 16 rounds to reach nearly 85\% pass rate before using ICL, while after using ICL, it exceeded 95\% in just three iterations. This proves that ICL can enhance the performance ceiling of role-playing. However, the gain from ICL is limited for the GPT series models. Specifically, GPT-4 and GPT-4o reach around 95\% pass rate within 1-2 iterations, indicating that the GPT series models already have a strong capability foundation for role-playing.

\section{Conclusion} 
We explored a personalized opinion summarization generation method based on LLMs and collected a test set to evaluate the level of personalization in the generated summaries using LLMs-based metrics.
In the design of the personalized opinion summarization method, we adopted a multi-agent framework where personalized summaries are generated through the interaction between the summarization system and the user system. To help the user system better understand the user's interests, role-playing was introduced into the user system, along with supervision and practice.
Experimental results demonstrate the effectiveness of this method.

\section{Limitation} 

Although experiments have shown that having the LLMs play the role of the user in the instructions helps it better understand user needs, user role-playing still slightly lags behind RAG in improving summarization performance. In the relevant experiments, it was observed that for GPT-4, the user role-playing pass rate reached over 90\% in the initial iterations, significantly higher than the other four models. This suggests that the model is unlikely to gain general improvements from the supervisor. However, in terms of summary performance, the results for GPT-4 were lower than those for GPT-4o and Qwen. This may indicate that although the model has a higher user role-playing pass rate, it has not captured truly valuable user information, or it may suggest that the model is unable to fully utilize the personalized information captured during summary generation. We hypothesize that the quality of the suggestions obtained through user role-playing may be limited by the supervisor's evaluation capability, and there may be room for improvement in how the summarization model utilizes user suggestions.

During the experiments, we also found that although the LLMs-based evaluation metrics can assess the personalization capability of the summaries without personalized summary labels, they often result in overly lenient ratings. While we made some adjustments to the evaluation metrics, alleviating this issue to a certain extent, we still found that the overall evaluation metrics tended to score too high in small-scale manual evaluations. This issue also occurred in the supervisor’s evaluation of user role-playing. More reasonable evaluation metrics or more controllable supervisor methods based on LLMs will be a key focus in our future work.


\bibliography{custom}

\appendix

\section{Dataset}
\label{dataset}

\begingroup
\renewcommand{\arraystretch}{1.10}
\begin{table}[!t]
\centering
\scalebox{0.83}{
\begin{tabular}{c|cccc}
\hline 
& $\textbf{High*2}$  & $\textbf{Low*2}$ & $\textbf{HHPL}$ \\
\hline Clothing & 100 & 50 & 46 \\
 Elec & 100 & 24 & 52 \\
 Home & 100 & 32 & 27 \\
 Health & 100 & 21 & 14 \\
\hline
\end{tabular}}
\caption{Analysis of the PerSum dataset. HHPL refers to samples with a high historical score but a low product score. 
}
\label{data_app}
\vspace{-0.2cm}
\end{table}
\endgroup

Since the sentiment in personalized reviews does not necessarily align with the overall sentiment of historical reviews. When evaluating the historical score, only aspect coverage is considered, excluding sentiment consistency. The quantity of samples in each category is displayed in Table \ref{data_app}. Due to the implementation of two filtering processes, the majority of the samples possess both high historical ratings and high product ratings. Consequently, the number of high-rating samples is randomly selected, with one hundred entries per category. The number of samples with low ratings is relatively small after filtering, which are included. 

\section{Implementation Details}
\label{detail}

During the experience, the temperature coefficient and other hyperparameters for all models were set to their default settings at the time of access, with no modifications. In cases where some models encountered input length limitations and could not process certain steps, the output for those steps was recorded as 'None'.

In the role-playing suggestion process, whether during the training or formal role-playing stages, when the supervisor generates “No suggestions,” it is considered that the current generated suggestion has passed the check. The iteration stops when one of the following three conditions is met. (1) When the iteration count does not exceed 5 and the pass rate reaches over 95\%. (2) when the iteration count exceeds 5 and the pass rate reaches over 85\%. (3) when the iteration count exceeds 15. Three practice rounds will be performed before generating the role-play.

In the Retrieval-augmented Rewriting process, multiple irrelevant reviews are removed at once for extraction efficiency.The extraction quantity is calculated as $(length_{review} - 10,000) / length_{review,avg}$. In cases where the review set exceeds the model's length limit, the review set is randomly divided into two parts for two extractions. 
If the target number of reviews cannot be obtained (due to the model generating too few, too many, or wrong numbers greater than the input text length), retries will be attempted. If the number of retries exceeds 5 and the target is still not reached, the process continues with additional retries until the goal is achieved or the retry threshold (set to 8) is reached. If the target is still not met after reaching the retry threshold, the remaining reviews will be randomly selected until the requirements are met. 
For historical reviews, if too many reviews are generated, a retry will be performed. If the retry count reaches 5, the first generated number in the next set will be selected as the target. During the final revision process, multiple modifications will be made using user suggestions, set to 5 iterations, and the highest result will be recorded to ensure that the suggestions are fully utilized.

\onecolumn

\section{Supervisor Prompt}
\label{super_prompt}
\begin{framed}
You are a ChatGPT role review expert, skilled in identifying and correcting any anomalous text in dialogue that may not align with the user's personality. Your goal is to evaluate whether the user's response is consistent with the their previous behavior based on historical comments, and to offer improvement suggestions. 

The suggested content should include a brief reason and specific, detailed revision advice. IF THERE ARE NO ERRORS OR SUGGESTIONS, you must write ONLY "No suggestions." in the suggestions section, without any explanation or additional words.
\newline

\# Attention

1. The user's response aims to assess whether the product review summary they have seen addresses the aspects they are interested in, and to provide suggestions for modifying the current summary.

2. The user's response should first summarize the product information they care about, then describe the relevant information in the current summary that matches their interests, and finally explain what additional information they would like to see included or what should be omitted.

3. The user's response should maintain role consistency while meeting basic readability and fluency requirements in the dialogue.
\newline

\# Metric

The consistency of a user's response includes historical exposure rate, accuracy, hallucination rate, and personal utterance consistency. 

1. The historical exposure rate refers to how much information from the user’s past comments is included in the response. 

2. Accuracy refers to whether the information presented about the user's is correct. 

3. The hallucination rate indicates whether the response includes information that doesn't belong to the user's. 

4. Personal utterance consistency refers to whether the content of the response aligns with the user’s personality and language style.
\newline

\# Steps

You will check the user's response through the following steps:

1. First, you should understand and analyze the user\'s response, the product summary, and the user's historical reviews.

1.1 For the input user's response, analyze the user’s understanding of both themselves and the summary, as well as the modification suggestions they offer.

1.2 For the input product summary, analyze the described product, its related aspects, and the emotional tone conveyed.

1.3 Based on the input user's historical reviews, analyze the user's personality traits, expression habits, shopping behavior, and aspects of products they are interested in.

2. Based on your analysis, check the consistency of the user’s response with the product summary and historical reviews, including historical exposure rate, accuracy, hallucination rate, and Personal utterance consistency.

2.1 For the historical exposure rate check, focus on how the user's response mentions product information they care about or are interested in. Then, based on other parts of the response, judge whether the user has clearly and concisely introduced their preferences, making subsequent requests for adding or reducing information in the summary reasonable and easy to understand.

2.2 For the accuracy check, pay attention to whether the personal information mentioned in the user’s response is consistent with their historical information, and whether the summary information matches the provided product summary.

\end{framed}
\begin{framed}

2.3 For the hallucination rate check, focus on whether there is any personal information mentioned in the user’s response that was not present in their historical comments, or whether there is product information in the response that was not mentioned in the summary.

2.4 For the personal utterance consistency check, pay attention to whether the tone of the user’s response is consistent with their personality, and whether the grammar and expression align with their usual communication style.

3. Output a summary of the above check results. If you find any errors or have any suggestions, clearly state them in the suggestions section. The suggested content should include a brief reason and specific, detailed revision advice. IF THERE ARE NO ERROR OR SUGGESTIONS, you must write ONLY "No suggestions" in the suggestions section.
\newline

\# Format example

Your final output should ALWAYS in the following format:
 
\#\# Thought

you should always think about if there are any errors or suggestions for user's response, NOT FOR SUMMARY.

\#\# Suggestions

1. ERROR1/SUGGESTION1

2. ERROR2/SUGGESTION2

2. ERROR3/SUGGESTION3
\newline

\# User's Response
\{ \}
\newline

\# Summary
\{ \}
\newline

\# User History Reviews
\{ \}
\newline

\end{framed}

\section{User Prompt}
\label{user_prompt}
The following demonstrates the instructions used by the user for generating and revising suggestions through role-playing.

\begin{framed}
Your task is to act as the user based on their historical reviews, evaluating whether the current summary addresses aspects you are interested in, and providing suggestions for modifications to the summary of the current product review. 
\newline

Your output format should be: 

\#\# Response:

\{

"Analysis": ANALYSIS

"Suggestions": SUGGESTIONS

\}
\newline

Here are some examples for you.

\{ \}
\newline

\# Previous reviews: \{ \}
\newline

\# Summary: \{ \}

\end{framed}

\begin{framed}
Your task is to act as the user based on previous reviews to evaluate whether the current summary addresses aspects you are interested in. However, there may have been some issues with your previous response, so please revise it based on the expert’s recommendations.
\newline

The response consists of two parts: the analysis and the suggestions. 

In the Analysis part, you should first briefly introduce yourself based on previous reviews, including aspects of the product that may interest you as well as those that do not. Then, identify which aspects of the current summary align with your interests, and what additional product details you would like to see. Additionally, you should point out which parts of the current summary are of no interest to you.

In the Suggestions part, you should provide clear and concise revision suggestions regarding what aspects should be added or reduced in the description .
\newline

ATTENTION: You are acting as the user, so you should use the first person for analysis and suggestions.

Your output format should be: 

\#\# Response:

\{

    \"Analysis\": ANALYSIS
    
    \"Suggestions\": SUGGESTIONS
    
\}

\# Previous reviews:\{ \}

\# Summary:\{ \}

\# Previous Response:\{ \}

\# Expert Recommendations:\{ \}

\end{framed}

\section{RAG Prompt}
\label{rag-prompt}

\begin{framed}
Please evaluate the relevance of the following reviews and provided text modification suggestions. 

Our goal is to identify reviews related to the current suggestions from the past reviews. 

Output the most relevant review number without explanation. ONLY ONE NUMBER IS NEEDED, and output format should be.

\{

"Numbers": [NUMBER]

\}

Suggestions:\{ \}

Review:\{ \}

\end{framed}

\section{Evaluation Prompt}
\label{eval-metric}

The original prompt for aspect coverage is the following. 

\begin{framed}
Task Description:

You will be given a set of reviews using which a summary has been generated. Your task is to evaluate the summary based on the given metric. Evaluate to which extent does the summary follows the given metric considering the reviews as the input. Use the following evaluation criteria to judge the extent to which the metric is followed. Make sure you understand the task and the following evaluation metric very clearly.
\newline

Evaluation Criteria:

The task is to judge the extent to which the metric is followed by the summary.

Following are the scores and the evaluation criteria according to which scores must be assigned.

<score>1</score> - The metric is not followed at all while generating the summary from the reviews

<score>2</score> - The metric is followed only to a limited extent while generating the summary from the reviews

<score>3</score> - The metric is followed to a good extent while generating the summary from the reviews.

<score>4</score> - The metric is followed mostly while generating the summary from the reviews.

<score>5</score> - The metric is followed completely while generating the summary from the reviews.
\newline

Metric:

Aspect Coverage - The summary should cover all the aspects that are majorly being discussed in the reviews. Summaries should be penalized if they miss out on an aspect that was majorly being discussed in the reviews and awarded if it covers all.
\newline

Reviews: 

\{\}

Summary: 

\{\}
\newline

Evaluation Steps:

Follow the following steps strictly while giving the response:

1. First write down the steps that are needed to evaluate the summary as per the metric. Reiterate what metric you will be using to evaluate the summary.

2. Give a step-by-step explanation if the summary adheres to the metric considering the reviews as the input. Stick to the metric only for evaluation.  

3. Next, evaluate the extent to which the metric is followed.

3. Use the previous information to rate the summary using the evaluation criteria and assign a score within the <score></score> tags.
\newline

Note: Strictly give the score within <score></score> tags only e.g Score- <score>5</score>.

First give a detailed explanation and then finally give a single score following the format: Score- <score>5</score>
\newline

THE EVALUATION AND SCORE MUST BE ASSIGNED STRICTLY ACCORDING TO THE METRIC ONLY AND NOTHING ELSE!

Response: 

\end{framed}

The current prompt for aspect coverage is the following.
\begin{framed}
Task Description:

You will be given a set of reviews using which a summary has been generated. Your task is to evaluate the summary based on the given metric. Evaluate to which extent does the summary follows the given metric considering the reviews as the input. Use the following evaluation criteria to judge the extent to which the metric is followed. Make sure you understand the task and the following evaluation metric very clearly.
\newline

Evaluation Criteria:

The task is to judge the extent to which the metric is followed by the summary.

Following are the scores and the evaluation criteria according to which scores must be assigned.

<score>0</score> - The metric is not followed at all while generating the summary from the reviews.

<score>20</score> - The metric is followed only to a limited extent while generating the summary from the reviews.

<score>40</score> - The metric is partially followed while generating the summary, but there are noticeable deficiencies.

<score>60</score> - The metric is followed to some extent while generating the summary, but there are several areas that require improvement.

<score>80</score> - The metric is followed mostly while generating the summary from the reviews.

<score>100</score> - The metric is followed completely while generating the summary from the reviews.
\newline

Metric:

Aspect Coverage - The summary should cover all the aspects that are majorly being discussed in the reviews. Summaries should be penalized if they miss out on an aspect that was majorly being discussed in the reviews and awarded if it covers all.
\newline

Reviews: 

\{\}

Summary: 

\{\}
\newline

Evaluation Steps:

Follow the following steps strictly while giving the response:

1. First write down the steps that are needed to evaluate the summary as per the metric. Reiterate what metric you will be using to evaluate the summary.

2. Give a step-by-step explanation if the summary adheres to the metric considering the reviews as the input. Stick to the metric only for evaluation.  

3. Next, evaluate the extent to which the metric is followed.

4. Use the previous information to rate the summary using the evaluation criteria and assign a score within the <score></score> tags.
\newline

Note: Strictly give the score within <score></score> tags only e.g Score- <score>100</score>.

First give a detailed explanation and then finally give a single score following the format: Score- <score>100</score>

THE EVALUATION AND SCORE MUST BE ASSIGNED STRICTLY ACCORDING TO THE METRIC ONLY AND NOTHING ELSE!
\newline

Response: 
\end{framed}

\end{document}